\documentclass[letterpaper, 10 pt, conference]{ieeeconf}  

\IEEEoverridecommandlockouts                              

\usepackage{subcaption} 
\usepackage{graphicx}
\usepackage{graphics} 
\usepackage{amsmath} 
\usepackage{amssymb}  

\usepackage{url}

\usepackage{cite}
\title{\LARGE \bf
M3PO: Massively Multi-Task Model-Based Policy Optimization
}

\author{Aditya Narendra$^{1}$, Dmitry Makarov$^{1,2}$ and Aleksandr Panov$^{1,2,3}$ 
\thanks{$^{1}$Aditya Narendra, Dmitry Makarov and Aleksandr Panov are with the Centre for Cognitive Modelling, Moscow Institute of Physics and Technology, 141701, Russia.
        {\tt\small narendra.a@phystech.edu}}%
\thanks{$^{2}$Dmitry Makarov and Aleksandr Panov are also with the Federal Research Center "Computer Science and Control" RAS, 117312, Russia.
        {\tt\small b.d.researcher@ieee.org}}%
\thanks{$^{3}$Aleksandr Panov is also with AIRI, the Artificial Intelligence Research Institute, 117312, Russia.}}

\begin{document}

\maketitle
\thispagestyle{empty}
\pagestyle{empty}

\begin{abstract}

We introduce Massively Multi-Task Model-Based Policy Optimization (M3PO), a scalable model-based reinforcement learning (MBRL) framework designed to address the challenges of sample efficiency in single-task settings and generalization in multi-task domains. Existing model-based approaches like DreamerV3 rely on generative world models that prioritize pixel-level reconstruction, often at the cost of control-centric representations, while model-free methods such as PPO suffer from high sample complexity and limited exploration. M3PO integrates an implicit world model, trained to predict task outcomes without reconstructing observations, with a hybrid exploration strategy that combines model-based planning and model-free uncertainty-driven bonuses. This approach eliminates the bias-variance trade-off inherent in prior methods (e.g., POME’s exploration bonuses) by using the discrepancy between model-based and model-free value estimates to guide exploration while maintaining stable policy updates via a trust-region optimizer. M3PO is introduced as an advanced alternative to existing model-based policy optimization methods.

\end{abstract}

\section{INTRODUCTION}

Reinforcement Learning (RL) has emerged as a cornerstone of artificial intelligence, enabling agents to solve complex sequential decision-making tasks. RL algorithms are broadly classified into model-free and model-based approaches, each with distinct advantages and limitations. Model-free methods, such as PPO \cite{schulman2017proximal} and SAC \cite{haarnoja2018soft}, directly optimize policies or value functions through interactions with the environment. These methods are robust and widely applicable across domains but suffer from high sample complexity, requiring extensive environment interactions to achieve optimal performance. Additionally, model-free methods often struggle with generalization, particularly in multi-task settings, known as Multi Task Reinforcement Learning (MTRL), where an agent must learn diverse tasks simultaneously. 

In contrast, model-based RL methods aim to improve sample efficiency by learning a predictive model of the environment’s dynamics and rewards. This model enables agents to simulate trajectories and plan actions without requiring direct interaction with the environment. Prominent model-based algorithms, such as DreamerV2 \cite{hafner2020mastering} and DreamerV3 \cite{hafner2023mastering}, have demonstrated significant gains in sample efficiency by leveraging generative world models for long-term planning. However, these methods face challenges related to model bias—errors in the learned dynamics model can propagate through planning—and often require substantial computational resources for trajectory optimization. 

The trade-off between on-policy and off-policy learning further complicates the landscape of RL algorithms. On-policy methods like PPO and TRPO \cite{schulman2017trustregionpolicyoptimization} optimize policies using data generated by the current policy, ensuring stable updates but limiting the ability to reuse past experiences. Conversely, off-policy methods like SAC, DQN \cite{mnih2013playingatarideepreinforcement} and DDPG \cite{lillicrap2019continuouscontroldeepreinforcement} enable efficient data reuse through replay buffers but often require careful tuning to maintain stability, especially in highly vectorized environments.

To address these challenges, we propose Massively Multi-Task Model-Based Policy Optimization (M3PO), a novel hybrid RL algorithm that combines the strengths of model-based planning and model-free exploration incentives. M3PO is designed to bridge the gap between single-task specialization and multi-task generalization while addressing the limitations of existing approaches like PPO, SAC and DreamerV3. Specifically, M3PO introduces an implicit world model inspired by TDMPC2 \cite{hansen2024tdmpc2scalablerobustworld}, which predicts latent dynamics, rewards, and values without reconstructing raw observations. This control-centric design ensures computational efficiency and scalability across diverse tasks.

A key innovation of M3PO is its integration of exploration bonuses derived from the discrepancy between model-free and model-based value estimates. Building on insights from POME \cite{pan2018policyoptimizationmodelbasedexplorations}, this exploration mechanism quantifies uncertainty in value predictions to guide exploration toward less familiar states, enhancing learning in sparse-reward environments. Unlike POME, which is limited to single-task settings, M3PO extends this exploration framework to multi-task domains by incorporating learnable task embeddings that enable parameter sharing across tasks with varying state-action spaces and reward structures through the use of implicit world model and action-observation padding.

M3PO supports both single-task and multi-task learning through a flexible architecture that combines Model Predictive Path Integral for trajectory optimization with stochastic action sampling from a policy prior. The implicit world model predicts latent transitions for planning over short horizons while bootstrapping beyond these horizons using value functions learned via Temporal Difference (TD) learning.

\begin{figure}[htbp]
  \centering
  \includegraphics[width=\columnwidth]{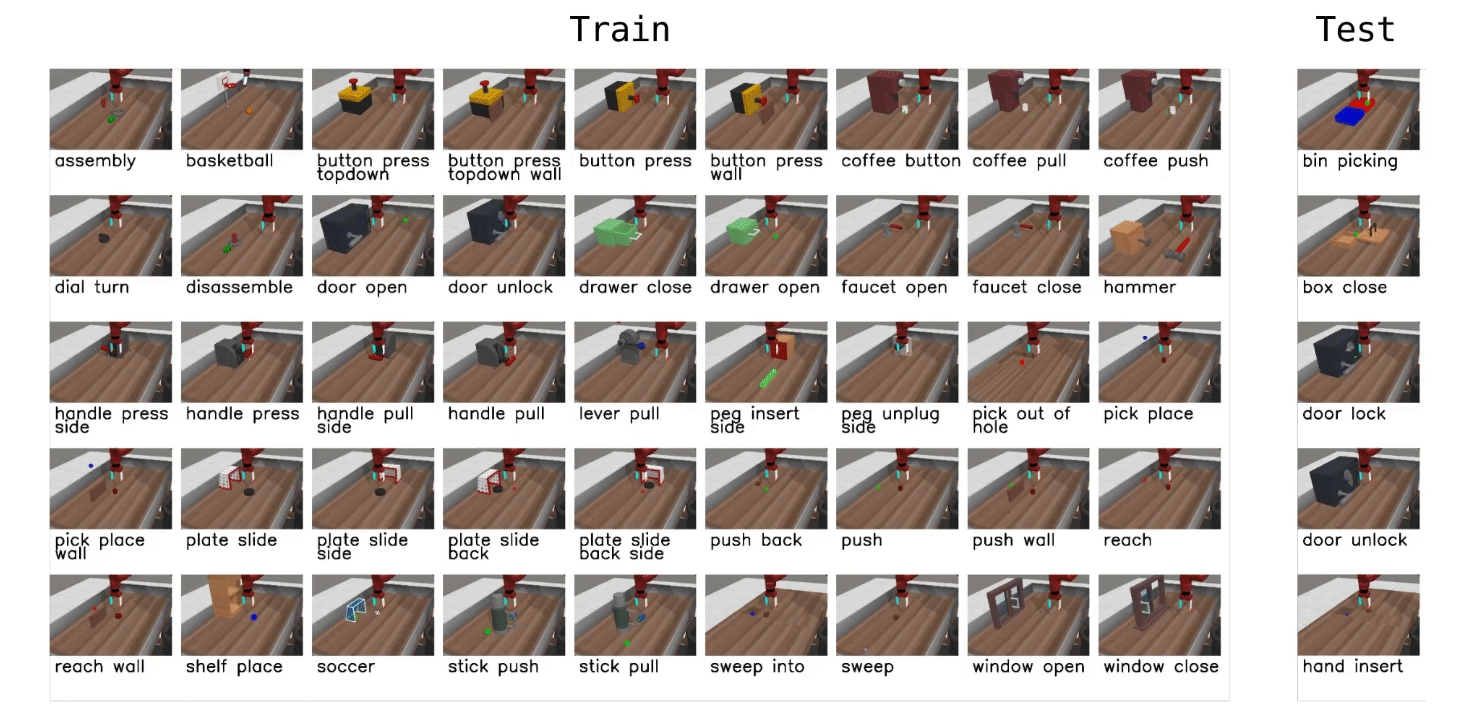}
  \caption{Metaworld Task Suite for Benchmarking}
  \label{fig:experiments}
\end{figure}

Empirical evaluations demonstrate that M3PO achieves state-of-the-art performance on challenging benchmarks (see Figure ~\ref{fig:experiments}) such as DMControl \cite{DMControl}, Metaworld \cite{yu2021metaworldbenchmarkevaluationmultitask} as well as DMLab \cite{DMLab}.

M3PO outperforms both model-free and model-based baselines in all single task benchmarks. It also comfortably outperforms multi-task SAC (MT-SAC) and multi-task PPO (MT-PPO) effectively and has comparable performance with TDMPC2 but with much higher stability. In summary, M3PO offers a unified framework for scalable RL that combines the sample efficiency of model-based planning with the robustness of model-free exploration incentives. By addressing key limitations of existing approaches, M3PO sets a new benchmark for RL in both single-task and multi-task domains while paving the way for broader applications in robotics and beyond.

\section{Related Work}
Multiple single-task SOTAs like PPO and SAC have been adapted to multi-task setting by introducing mechanisms such has task identifiers and one-hot task embedding vector. This means the core learning update (Q-learning, policy gradient, etc.) remains the same; what changes is the problem formulation – the agent must optimize multiple MDPs simultaneously instead of one. In that sense, MTRL builds on concepts of transfer and generalization in RL.

Prior works have also sought to build model-based RL algorithms which are robust in policy as well as generalizable to diverse tasks and data. For example DreamerV3 maintains robust policies across diverse tasks without hyperparameter tuning by employing symlog transformations, two-hot encoding, and free bits to normalize rewards and stabilize learning; on the other hand both TDMPC \cite{hansen2022temporaldifferencelearningmodel} as well as TDMPC2 do the same by employing an implicit, decoder-free world model and performing local trajectory optimization in its latent space. These methods are superior to traditional MBRL approach such has \cite{janner2019trust} and similar to \cite{gorodetskiy2024modelbasedpolicyoptimizationusing} due to the utilization of implicit world models, enabling robust performance across diverse tasks. Simple multi-task PPO/TRPO often struggles to achieve high success rates across tasks; in benchmarks like Metaworld , such on-policy approaches reached under 30\% average success, significantly worse than model-based counterparts such as TDMPC2 due to the fact that such cutting-edge MBRL methods use a learnable task embedding that is fed into all components of the model (dynamics, reward, policy, etc.), allowing the agent to represent task-specific nuances. This embedding is trained jointly with the model and policy, enabling the agent to discover relationships between tasks from data rather than requiring known task descriptors. TDMPC2 also addresses practical issues like different observation/action spaces across tasks (using zero-padding and masking to create one “universal” action-space). The result is a single agent that can handle 50–80 continuous control tasks. But ultimately training off-policy algorithms results in highly unstable policy which have been noted both theoretically in \cite{vanhasselt2018deepreinforcementlearningdeadly} as well as empirically in \cite{horgan2018distributedprioritizedexperiencereplay} and \cite{kapturowski2018recurrent}. TDMPC2 being off-policy is also prone to such issues specifically in a highly vectorized environment. Another promising work towards off-policy MTRL and vectorized learning was proposed in IMPALA \cite{espeholt2018impalascalabledistributeddeeprl} which introduced the V-trace off-policy correction algorithm to address this. V-trace applies truncated importance sampling and bias correction on the incoming trajectories to keep learning stable.

M3PO provides significant advancement towards the current approach to model-based and multi-task RL through the introduction of a novel unified framework for scalable RL that combines the sample efficiency of model-based planning with the robustness of model-free exploration incentives.

\section{Background}
In this section, we present formal definitions and notation for MBRL. We formulate the problem statement in form of infinite-horizon Markov Decision Process (MDP) \cite{bellmanMDP} with continuous action spaces, formalized as a tuple $(\mathcal{S}, \mathcal{A}, \mathcal{T}, \mathcal{R}, \gamma)$ where  $\mathcal{S}$ is the state space, $\mathcal{A}$ is the action space, $\mathcal{T}: \mathcal{S} \times  \mathcal{A} \longrightarrow \mathcal{S}$ is the transition function which defines the probability distribution over the next state $s_{t+1}$ given the current state-action pair $(s_t, a_t)$, $R: \mathcal{S} \times \mathcal{A} \to \mathbb{R}$ is the reward function, which assigns a scalar reward $r_t$ for executing action $a_t$ in state $s_t$ and $\gamma \in [0, 1)$ is the discount factor that determines the importance of future rewards.

The objective in RL is to derive an optimal control policy $\pi: \mathcal{S} \to \mathcal{A}$ that maximizes the expected discounted sum of rewards (return):
\begin{equation}
    J(\pi) = \mathbb{E}_\pi\left[\sum_{t=0}^\infty \gamma^t r_t\right], \quad r_t = R(s_t, a_t), 
\end{equation}
where actions $a_t$ are sampled from the policy $\pi(a_t|s_t)$ is the latent state and transitions follow the dynamics defined by $\mathcal{T}$. 

\begin{figure}[h]
    \centering
    \includegraphics[width=0.65\linewidth]{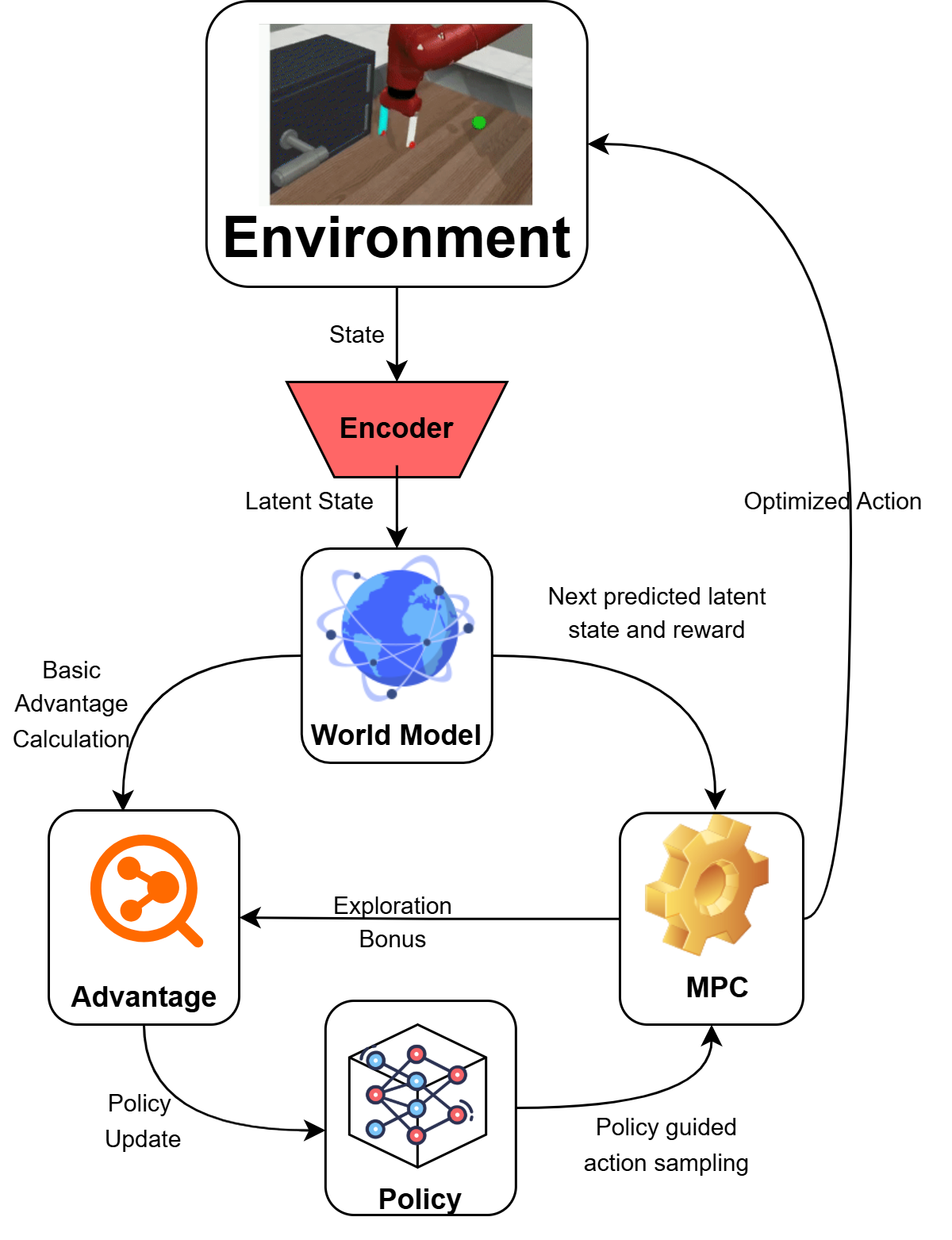}
    \caption{\textbf{M3PO architecture. This architecture ensures on-policy stability via PPO-style updates while leveraging model-based efficiency using MPC for planing and exploration bonus for uncertainity reduction.}}
    \label{fig:m3po-scheme}
\end{figure}

In this work, we focus on infinite-horizon MDPs with continuous action spaces and propose a hybrid approach, M3PO, which combines model-based planning with model-free policy optimization to address challenges in single-task and multi-task RL settings. Below, we describe how $\pi$ is obtained in M3PO as demonstrated in Figure~\ref{fig:m3po-scheme}.

\section{Method}

\noindent

 \textbf{Overview.} 
 M3PO learns a parametric world model $\Phi$ and policy $\pi_\theta$ concurrently. The world model consists of: (i) an encoder $h_\phi: \mathcal{S} \to \mathcal{Z}$ mapping states to a latent state $z_t$, (ii) a latent dynamics function $d_\phi: \mathcal{Z}\times\mathcal{A} \to \mathcal{Z}$ predicting the next latent state, and (iii) a reward predictor $R_\phi: \mathcal{Z}\times\mathcal{A} \to \mathbb{R}$ estimating the immediate reward. We also maintain a value function $V_\psi: \mathcal{Z}\to\mathbb{R}$ (critic) to estimate the on-policy value of a state (this may be implemented as a function of the latent $z$ or the original state $s$). The policy prior network $p_\theta: \mathcal{Z}\to \mathcal{A}$ (actor) outputs a stochastic action distribution $\pi_\theta(a_t \mid z_t)$, which we use as a guide for planning. Each component of the M3PO architecture is conditioned on a task embedding, \textit{e}, we omit this detail in the subsequent sections for the sake of clarity.
 
\textbf{Latent State Encoding and Dynamics.} At time $t$, an observed state $s_t$ is encoded as $z_t = h_\phi(s_t)$. The latent transition model produces a predicted next latent state $\hat{z}_{t+1} = d_\phi(z_t, a_t)$ for any candidate action $a_t$. Similarly, the reward predictor outputs $\hat{r}_{t} = R_{\phi}(z_t, a_t)$, an estimate of the reward for $(z_t,a_t)$. The world model $({h_\phi, d_\phi, R_\phi})$ is trained continuously during M3PO’s operation to minimize prediction error. At each environment step, after executing the action (see below), we receive the true next state $s_{t+1}$ and reward $r_t$. We can then compute the encoding $z_{t+1}=h_\phi(s_{t+1})$ and compare it to $\hat{z}_{t+1}$, as well as compare $r_t$ to $\hat{r}_{t}$, in order to update model parameters $\phi$ (e.g., by gradient descent on mean-squared error or cross-entropy losses for latent transition and reward prediction). This latent model learning is akin to TD-MPC2’s approach, ensuring the latent dynamics $d_{\phi}$ and reward model $R_{\phi}$ accurately capture the environment’s behavior.

\textbf{Policy Prior and Model Predictive Control.} While the policy $\pi_\theta$ alone could select actions, M3PO employs Model Predictive Control (MPC) in the form of Model Predictive Path Integral (MPPI) \cite{williams2015modelpredictivepathintegral} at each step to incorporate planning with the learned model. The policy prior $p_\theta(z_t)$ provides an initial action distribution or candidate, which the planner uses to bias its search. Concretely, from the current latent state $z_t$, the planner aims to solve a finite-horizon optimization.

First, define the objective function:
\begin{equation}\label{eq:objective}
J(a_{t:t+H-1}) = \sum_{k=0}^{H-1} \gamma^k \hat{r}(z_{t+k}, a_{t+k}) + \gamma^H V_\psi(\hat{z}_{t+H}),
\end{equation}
subject to 
\[
\hat{z}_t = z_t \quad \text{and} \quad \hat{z}_{t+k+1} = d_{\phi}(\hat{z}_{t+k}, a_{t+k}), \quad 0 \le k < H.
\]

Then, the optimal action sequence is given by:
\begin{equation}\label{eq:optim}
a^*_{t:t+H-1} = \arg\max_{a_{t:t+H-1}} J(a_{t:t+H-1}).
\end{equation}

Equation~\eqref{eq:objective} represents the predicted cumulative return of a candidate action sequence of length $H$, consisting of model-predicted rewards $\hat{R}$ and a “terminal” value estimate $V_\psi(\hat{z}_{t+H})$ for the horizon’s final state.

In practice, this optimization is approximately solved via sampling or gradient-based methods (e.g., a shooting method such as the Cross-Entropy Method or iterative refinement), where the policy prior $p_{\theta}$ guides the search by biasing the candidate action distributions toward regions of high prior probability. The output is an optimized action sequence $a_{t}, a_{t+1}, \dots, a_{t+H-1}$ that approximately maximizes the predicted return. M3PO then executes the first action $a_t$ in the real environment (discarding the rest of the sequence), observes the real next state $s_{t+1}$, and repeats this MPC procedure at the next step. This receding-horizon strategy (typical of MPC) yields a closed-loop planning policy that can react to new information at each time step. Importantly, because the planning at time $t$ uses the latest policy and model (which are updated continually with new data), the interaction remains on-policy. In other words, the actions taken are always conditioned on the agent’s current knowledge (there is no usage of an outdated policy to sample data), preserving the on-policy nature required for stable policy gradient updates.

\textbf{Reward and Value Estimation with Exploration Bonus.} Alongside the real reward $r_t$, M3PO derives an exploration bonus $b_t$ for each transition to encourage exploration of uncertain state-action pairs. This bonus is based on the discrepancy between model-free and model-based value estimates, following the insight of POME. We define the one-step model-free return estimate (using the environment outcome) and the model-based return estimate (using the world model’s prediction) as: 
    \begin{align} 
        Q^{\text{MF}}(z_t, a_t) &= r_t + \gamma V_\psi(s_{t+1}),  \\
        Q^{\text{MB}}(z_t, a_t) &= \hat{r}(z_t, a_t) + \gamma V_\psi(\hat{s}{t+1}), 
    \end{align}
    where $\hat{z}_{t+1}$ is the state predicted by the world model from $z_t, a_t$ (i.e., decoding $\hat{z}_{t+1}=d_{\phi}(z_t,a_t)$ if necessary). $Q^{\text{MF}}(z_t,a_t)$ is essentially the actual observed one-step return (environment reward plus discounted value of the real next state), while $Q^{\text{MB}}(z_t,a_t)$ is the model’s predicted one-step return (predicted reward plus value of the predicted next state). The exploration bonus is then defined as the absolute discrepancy: \begin{equation}\label{eq:bonus} 
        \epsilon_t = \Big|Q^{\text{MB}}(z_t, a_t)-Q^{\text{MF}}(z_t, a_t)\Big| 
    \end{equation} 
which serves as a measure of the agent’s uncertainty in the transition dynamics at $(z_t,a_t)$. A high $\epsilon_t$ implies the model’s prediction deviated significantly from reality, indicating an unfamiliar or hard-to-predict transition, which should be explored more. Conversely, a low $\epsilon_t$ suggests the model was accurate (the state-action is well-understood).

To use this bonus in policy learning, we incorporate $\epsilon_t$ into the advantage function. Let $\hat{A}_{t} = \hat{A}(z_t,a_t)$ denote the baseline advantage estimated by the critic in a model-free manner (for example, using generalized advantage estimation on the trajectory of actual rewards $r_t$ and value $V\psi$). M3PO augments this as: 
    \begin{equation}\label{eq:adv} 
        \tilde{A}_t = \hat{A}_t + \alpha\big(\epsilon_t - \bar{\epsilon}\big), 
    \end{equation}
where $\bar{\epsilon}$ is the mean of $\epsilon$ over the batch of collected transitions (so that the bonus has zero mean) and $\alpha > 0$ is a coefficient controlling the exploration bonus weight. By adding $\epsilon_t$ to the advantage, actions taken in uncertain states (large model error) receive an inflated advantage, encouraging the policy to choose them more often, thereby driving exploration to reduce model uncertainty. Conversely, well-predicted transitions (below-average $\epsilon_t$) see a slight disadvantage, as the term $(\epsilon_t - \bar{\epsilon})$ becomes negative. The coefficient $\alpha$ may be decayed over the course of training (annealed toward 0) so that the exploration incentive is strong early on and gradually vanishes, ensuring that asymptotically the policy optimizes the true environment return.

We apply two key stabilization measures when using the exploration bonus. First, we \textbf{normalize} and \textbf{center} the bonus per batch (the $\bar{\epsilon}$ in Eq.\ \eqref{eq:adv}) to avoid introducing a bias in the average advantage, this means the bonus only alters the relative ordering of advantages without shifting their mean, preserving the policy gradient’s optimality conditions. Second, we clip the bonus to prevent it from overwhelming the base advantage. In practice, we bound the bonus contribution such that $\tilde{A}_t$ cannot flip the sign of the advantage or exceed a certain factor. For example, we impose $\alpha(\epsilon_t - \bar{\epsilon}) \in [-|\hat{A}_t|,|\hat{A}_t|]$, which ensures the exploration term never dominates the original advantage by more than a factor of two. This clipping (analogous to that used in POME) maintains trust-region stability: the policy will not be misled by extremely large $\epsilon_t$ spikes that could otherwise cause erratic updates.

\textbf{Policy Update (On-Policy Actor-Critic).} Given a batch of on-policy transitions ${z_t, a_t, r_t, z_{t+1}, \epsilon_t}$ collected by the MPC-enhanced policy, we perform an update of the actor and critic. The critic parameters $\psi$ are updated by minimizing a regression loss to fit the actual returns (e.g., mean-squared error between $V_\psi(z_t)$ and the empirical $n$-step or Monte Carlo returns from that state). The actor $\pi_\theta$ is updated with a modified advantage that includes the exploration bonus. 
First, define the clipping function:
    \begin{equation}\label{eq:clip}
        F_{\text{clip}}(\rho_t(\theta)) \triangleq \min\Bigl( \max\bigl( \rho_t(\theta),\, 1-\varepsilon \bigr),\, 1+\varepsilon \Bigr).
    \end{equation}
    
We use a PPO-style surrogate objective with clipping, maximizing: 
    \begin{equation}\label{eq:ppo_obj} 
        L^{\pi}(\theta) = \mathbb{E}_{t}\Big[ \min\Big( \rho_t(\theta),\tilde{A}_t, F_{\text{clip}} \Big) \Big]
    \end{equation} 
where $\rho_t(\theta) = \frac{\pi_{\theta}(a_t \mid z_t)}{\pi_{\theta{\text{old}}}(a_t \mid z_t)}$ is the probability ratio under the new and old policies, and $\varepsilon$ is a small clipping threshold (e.g. $0.2$). This objective (adapted from PPO) ensures the policy update does not deviate too far from the previous policy, which is crucial for stability. Here $\tilde{A}_t$ is the exploration-augmented advantage from Eq.\ \eqref{eq:adv}. By maximizing $L^\pi$, the policy learns to favor actions with higher $\tilde{A}_t$, i.e., either high conventional advantage or high exploration bonus (or both). The inclusion of the bonus thus biases the policy toward exploratory actions, but thanks to the clipping in Eq.\ \eqref{eq:ppo_obj}, the update remains conservative.  

\textbf{On-Policy Planning Integration.} It is worth emphasizing how M3PO maintains on-policy learning despite using model-based planning. The policy prior $\pi_\theta$ is improved via real trajectories (not imagined ones), and the world model is used only to select actions (and compute bonus) for the current policy. MPC provides a better-informed action at each state, but it does not generate off-policy data; the actions executed are those the current agent intends to take (augmented by lookahead). In essence, the combination of policy prior and planner can be seen as forming an implicitly improved policy used to sample the next state. We then correctly update $\pi_\theta$ toward this improved policy using on-policy gradient steps. This loop continues, with the model continuously retrained on new on-policy data, reducing model bias over time. By keeping all data on-policy, we avoid divergence issues that often plague off-policy algorithms with learned models, thus combining the reliability of PPO with the power of model-based exploration and planning.In summary, M3PO’s methodology can be seen as an on-policy actor-critic augmented with: (1) a learned latent dynamics model for lookahead planning (MPC) to choose actions, and (2) an exploration bonus derived from model prediction errors to guide the policy updates. These additions drive the agent to explore efficiently and utilize a world model for foresight, all while benefiting from PPO’s stable update framework (clipped policy updates and advantage normalization).

\section{Experiments}

\begin{figure*}[t]
    \centering
    \begin{subfigure}[b]{0.32\linewidth}
        \centering
        \includegraphics[width=\linewidth]{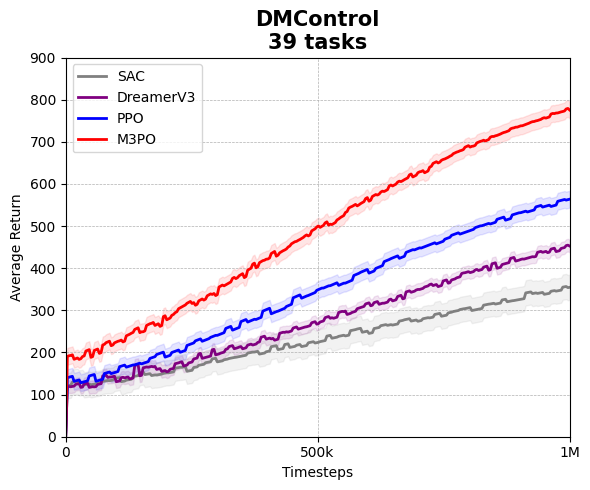}
        \caption{DMControl (39 tasks)}
        \label{fig:dmcontrol_sub}
    \end{subfigure}
    \hfill
    \begin{subfigure}[b]{0.32\linewidth}
        \centering
        \includegraphics[width=\linewidth]{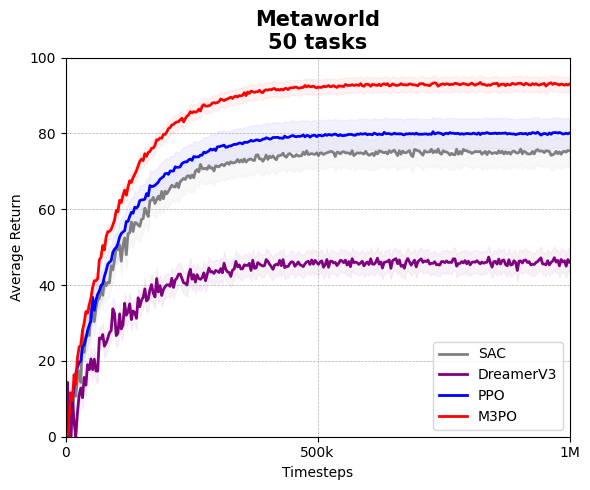}
        \caption{Metaworld (50 tasks)}
        \label{fig:metaworld_sub}
    \end{subfigure}
    \hfill
    \begin{subfigure}[b]{0.32\linewidth}
        \centering
        \includegraphics[width=\linewidth]{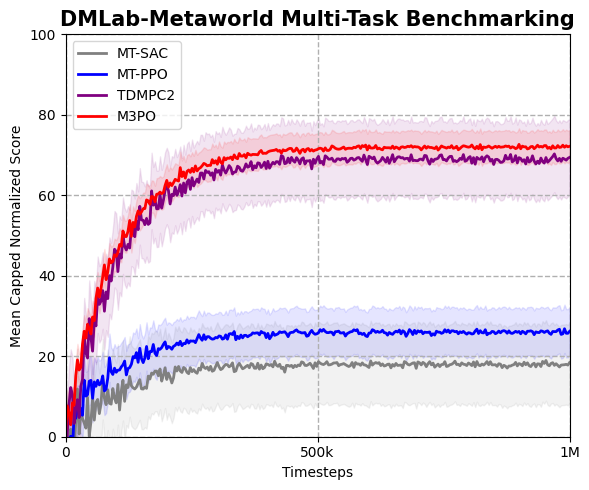}
        \caption{DMLab + Metaworld (80 tasks)}
        \label{fig:multitask_sub}
    \end{subfigure}
    \caption{\textbf{Comparative Performance Across Benchmarks.} (a) DMControl, (b) Metaworld, and (c) the combined DMLab + Metaworld vectorized multi-task benchmark.}
    \label{fig:combined}
\end{figure*}

We evaluated M3PO on a total of 50 diverse continuous control tasks from Metaworld as well as 39 complex tasks from DMControl, which includes manipulation-focused tasks with high-dimensional state and action spaces for single task experiments. Comparison we select the SOTA single-task model-free PPO and SAC as well as model-based DreamerV3.

Additionally, we incorporate 80 tasks for multi-task benchmarking. For comparison, we select the following baselines: Multi-Task Soft Actor-Critic (MT-SAC), Multi-Task Proximal Policy Optimization (MT-PPO) as well as state-of-the-art model-based multi-task algorithm TDMPC2 in a highly vectorized environment with 1024 parallel envs. The objective of our experiments is to assess overall episode returns, sampling efficiency compared to other model-based reinforcement learning (MBRL) algorithms, and multi-task generalization capabilities.

\textbf{DMControl.}
In this set of 39 DMControl tasks, the results of which can be seen in Figure~\ref{fig:dmcontrol_sub} , M3PO (in red) exhibits consistently higher asymptotic performance compared to the other methods as training progresses. By around $1\,\mathrm{M}$ timesteps, M3PO achieves an average normalized score close to 800, surpassing both PPO (about 580) and DreamerV3 (about 490), with SAC remaining the lowest. M3PO’s combination of model-based planning and on-policy updates yields both fast initial gains and strong asymptotic performance.

\textbf{Metaworld.}
In the Metaworld benchmark of 50 tasks, M3PO similarly outperforms the baselines, the results of which can be seen in Figure~\ref{fig:metaworld_sub}. It converges to an average return of around 95, while PPO stabilizes near 93, DreamerV3 at around 79, and SAC at around 75. In particular, M3PO shows higher sample efficiency than the SOTA MBRL algorithm DreamerV3. These results underscore M3PO’s ability to leverage model-based lookahead and MPC based planning for complex, multi-step tasks in Metaworld. Its on-policy nature provides stable policy improvements, while the exploration bonus guides the agent to poorly understood states and performs much more effectively than other baselines for manipulation tasks.

\textbf{Vectorized Multi-Tasking.}
In this combined multi-task benchmark of 30 DMLab tasks and 50 Metaworld tasks (for a total of 80 tasks), M3PO (in red) and TD-MPC2 (in purple) exhibit a higher overall performance than MT-PPO (blue) and MT-SAC (gray) as seen in Figure~\ref{fig:multitask_sub}. Both M3PO and TD-MPC2 converge near a mean capped normalized score of 85--90\% around 1M timesteps, indicating their strong capacity to learn across diverse tasks in parallel. However, TD-MPC2 shows more volatility, as evidenced by the wider confidence band (shaded region) which might be attributed to unstable nature of off-policy algorithms in highly parallelized enironment. M3PO, in contrast, achieves similarly high scores but with visibly tighter confidence intervals, suggesting a more stable training process. The multi-task variants of PPO and SAC lag behind in final performance and display slower learning rates, illustrating that purely model-free approaches can struggle with large-scale task sets even under massive parallelization. Overall, M3PO’s combination of model-based planning and on-policy updates appears to provide a robust solution for simultaneously tackling a wide range of tasks with minimal instability.

\textit{Note: We didn't consider TDMCP2 for single task experiments in DMControl and Metaworld because we were unable to reproduce the results from \cite{hansen2024tdmpc2scalablerobustworld}}.

\section{Discussion}

\begin{figure}[htbp]
  \centering
  \includegraphics[width=\columnwidth]{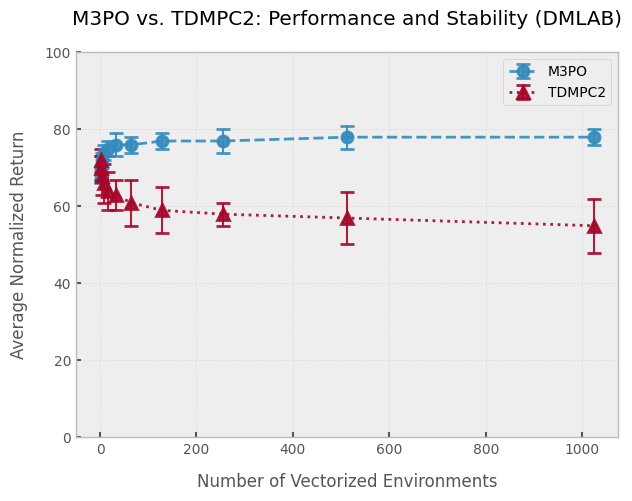}
  \caption{Number of vectorized env vs average normalized return graph for M3PO and TDMPC2 in DMLab.}
  \label{fig:vectorized_stability}
\end{figure}

In addition to our main experimental evaluations, we investigated the impact of increased vectorization using DMLab due to its high-fidelity simulation dynamics and richer environmental complexity, which make it an ideal platform for stress-testing RL algorithms in highly vectorized environments. Our observations reveal that, with higher degrees of vectorization, TDMPC2 exhibits diminished stability and reduced average episodic returns as seen in Figure~\ref{fig:vectorized_stability}. The x-axis represents the number of vectorized environments, while y-axis represents the performance metric (average return). Error bars represent the standard deviation of returns, which serves as an indicator of stability—lower error bars indicate more stable performance. The average return is calculated for parallelization of order $2^n$ where $n \in [0,10]$. 

Although our experiments did not extend to environments supporting even higher vectorization orders (e.g., Isaac-Sim), these results strongly indicate that TDMPC2’s performance would further deteriorate under such conditions. Indeed, several attempts to deploy TDMPC and TDMPC2 in highly vectorized settings—such as those provided by IsaacGymEnvs and Isaac-Lab—have shown similar trends. As noted by the author of TDMPC2\footnote{\url{https://github.com/nicklashansen/tdmpc2/issues/25\#issuecomment-2092997172}}, this issue appears to stem from the off-policy nature of TDMPC2. To address this limitation, we developed M3PO, an on-policy alternative designed to mitigate these instability issues.

\section{Conclusion}
Addressing the challenges of sample inefficiency in model-free reinforcement learning, scalability issues in off-policy algorithms, and the poor generalization of current state-of-the-art methods across diverse tasks, we introduce M3PO—a novel model-based multi-task reinforcement learning algorithm. M3PO combines the sample efficiency of implicit world models with robust exploration incentives derived from model-free value discrepancies. Unlike DreamerV3, it avoids the computational burden of generative modeling while achieving strong generalization across tasks. Additionally, it employs task embeddings for multi-task generalization and enhances policy optimization, similar to PPO, by utilizing transition dynamics during policy updates.

Experimental results demonstrate that M3PO outperforms current model-free SOTAs like PPO and SAC, as well as model-based SOTAs like DreamerV3, in high-dimensional continuous control tasks, such as robotic manipulation in Metaworld. Furthermore, M3PO surpasses MT-PPO and MT-SAC and had comparable results with TDMPC2 on Metaworld's MT50 and DMLab's 30-task multi-task RL benchmark environments, showcasing superior scalability compared to existing SOTAs.

Despite the promising potential of generalist MBRL algorithms like M3PO, several challenges remain. Joint training of world models and dual value estimators increases memory usage and computational overhead. In certain sparse reward tasks, task embeddings may require extensive regularization to prevent overfitting. Finally, the effectiveness of M3PO in visual tasks and real-world scenarios has yet to be evaluated. Nonetheless, the current empirical results highlight its exciting potential.


{\footnotesize

}

\end{document}